\documentclass{article}

 \PassOptionsToPackage{numbers}{natbib}
\usepackage[final]{nips_2017}

\usepackage[utf8]{inputenc} 
\usepackage[T1]{fontenc}    
\usepackage{hyperref}       
\usepackage{url}            
\usepackage{booktabs}       
\usepackage{amsfonts}       
\usepackage{nicefrac}       
\usepackage{microtype}      
\usepackage{amsmath,amsthm,amsfonts,amssymb,bm}
\usepackage{enumitem}
\usepackage{graphicx}
\usepackage{bm}
\usepackage{chngpage}
\usepackage{algorithm}
\usepackage{algorithmic}
\usepackage{subcaption}
\usepackage{caption}
\usepackage{color}
\usepackage{wrapfig}
\usepackage{bbm}

\newtheorem{Thm}{Theorem}

\newcommand{\vct}{\boldsymbol }

\DeclareMathOperator*{\argmax}{arg\,max}

\title{Racing Thompson: an Efficient Algorithm for Thompson Sampling with Non-conjugate Priors}

\author{
  Yichi Zhou$^\dag$,~~~  Jun Zhu$^\ddag$\footnotemark[1], ~~~ Jingwe Zhuo$^\dag$ \\
$^\dag$ Department of Computer Science \& Technology, TNList Lab, Tsinghua University, Beijing, 100084\\
$\ddag$ Department of Electronic Engineering, Tsinghua University, Beijing, 100084 \\ 
  \texttt{\{zhouyc15\}@mails.tsinghua.edu.cn;~dcszj@tsinghua.edu.cn} 
}

\begin{document}

\maketitle

\begin{abstract}Thompson sampling  has impressive empirical performance for many multi-armed bandit problems. But current algorithms for Thompson sampling only work for the case of conjugate priors since these algorithms require to infer the posterior, which is often computationally intractable when the prior is not conjugate. In this paper, we propose a novel algorithm for Thompson sampling which only requires to draw samples from a tractable distribution, so our algorithm is efficient even when the prior is non-conjugate. To do this, we reformulate Thompson sampling as an optimization problem via the Gumbel-Max trick. After that we construct a set of random variables and our goal is to identify the one with highest mean. Finally, we solve it with techniques in best arm identification. 
\end{abstract}

\section{Introduction}
\label{Sec:intro}
In multi-armed bandit (MAB) problems~\cite{lai1985asymptotically}, an agent chooses an action (in the literature of MAB, an action is also named as an arm.) from an action set repeatedly, and the environment returns a reward as a response to the chosen action. The agent's goal is to maximize the cumulative reward over a period of time. In MAB, a reward distribution is associated with each arm to characterize the uncertainty of the reward. 
One key issue for MAB and many on-line learning problems~\cite{bubeck2012regret} is to well-balance the exploitation-exploration tradeoff, that is, the tradeoff between choosing the action that has already yielded greatest rewards and the action that is relatively unexplored. 

As one of the most important problems in learning and decision-making in unknown environments, MAB has been studied in various settings since the seminal work \citep{lai1985asymptotically} (see \citep{bubeck2012regret} for a nice review). In this paper, we consider Bayesian bandits \citep{scott2010modern}, which is a well-studied variant of MAB. In a Bayesian bandit, the agent has a prior distribution on the mean of the reward distribution for each action .
The agent makes decisions adaptively according to the prior distributions and the past observations of each action. The most popular algorithm for Bayesian bandits is known as Thompson sampling (TS), which has a long history tracing back to \citep{thompson1933likelihood}. TS has proven to be powerful in practice \citep{chapelle2011empirical} with theoretical guarantees \citep{agrawal2012analysis,kaufmann2012thompson}. 

TS selects each arm randomly according to its probability to be optimal given the previous observations. Existing implementation of TS requires to infer the posterior distribution (See Section \ref{sec:preliminaries} for details), which can be computationally intractable in sophisticated models; thereby limiting the scope of Bayesian bandits to use simple conjugate priors. 
However, non-conjugate prior is very important in MAP. On one hand, non-conjugacy naturally arises because of using either a flexible prior or a flexible observation model (i.e., likelihood) to characterize the complex properties often appearing real-world applications. For example, in the Web Advertising task proposed by \citep{gopalan2014thompson}, it is natural to use a Bernoulli distribution to model the binary events whether a user clicks or not, and the prior distributions of different advertises are not independent because of the similarity between two advertises or other factors. In this case, it is not likely to have a conjugate prior that well incorporates such dependence. 
Another example is in \citep{kawale2015efficient}, where TS is used to do matrix-factorization recommendation in an online manner and their observation is a product of two Gaussian random variables with zero mean. An inverse gamma distribution is used as the prior on their variances, which is non-conjugate with the observation model. 
On the other hand, 
TS with an improper prior distribution can lead to worse performance that that with a correct prior, which can be non-conjugate. For example, in the well-known stochastic Bernoulli bandits problems \citep{lai1985asymptotically}, we have no prior knowledge on the mean of these arms. In this case, \citep{kaufmann2012thompson} shows that TS with a uniform prior achieves asymptotically optimal performance that matches the lower bound proved by \citep{lai1985asymptotically}, while the best known theoretical result by \citep{agrawal2012analysis} with a conjugate Beta prior can lead to worse performance. Moreover, as we shall see in Section \ref{sec:exp}, an incorrect prior indeed yields worse performance in practice.

As the exact posteriors are often intractable in the non-conjugate case, \citep{gopalan2014thompson} attempts to approximate the posteriors via sequential Monte-Carlo (SMC) \citep{del2014particle}. SMC maintains a set of particles and resamples sequentially according to the observations. However, to the best of our knowledge, there is no standard method to select the number of particles to maintain. Maintaining a large number of particles will cause inefficient computation, while a small number can be inaccurate. Besides, while Markov Chain Monte Carlo (MCMC) is widely use to draw posterior samples, its computational cost of sampling $p(\mu|X_1,...,X_t)$ increases with time $t$ \cite{cherkassky2013sequential}. Overall, the above problems restrict the applications of TS significantly.

\textbf{Contributions}: In this paper, we present a novel racing algorithm to implement TS, which can apply to the general case with non-conjugate priors. The algorithm approximates Thompson sampling, and only requires to sample from tractable distributions while avoiding posterior inference. Thus, the proposed algorithm is efficient. Our algorithm also needs to draw a set of samples similar to SMC, but our algorithm has a simple guideline to determine the number of samples with theoretical guarantees. Furthermore, our algorithm works well in experiments.

Technically, our method is built on a novel reformulation of TS as an optimization problem by exploring the Gumbel-Max trick~\citep{papandreou2011perturb}.
The goal of the optimization problem is to find the variable with the maximum expectation among a set of variables. Such a problem reduces to a best arm identification (BAI) problem \citep{kaufmann2015complexity,maron1997racing,kalyanakrishnan2012pac,jamieson2014best}, which is a well-studied variant of stochastic multi-armed bandits. To compute the expectations as involved in the BAI problem, we can freely construct a tractable prior for easy and efficient sampling, therefore avoiding the posterior inference with a non-conjugate prior.


The rest of the paper is structured as follows. We discuss some related work in Section~\ref{sec:related-work}. Section~\ref{sec:preliminaries} reviews some preliminary knowledge of TS.  Then, we present our algorithm with the new reformulation in Section~\ref{sec:algorithm}. Empirical studies are presented in Section~\ref{sec:exp}. Finally, we conclude in Section~\ref{sec:conclusions}.

\section{Related work}\label{sec:related-work}
In this section, we briefly summarize the related work in Bayesian bandits, the Gumbel-Max trick and the best arm identification. 

\textbf{Bayesian bandits}: Bayesian bandits has a long history dates back to \citep{thompson1933likelihood}, in which TS is introduced. TS is a kind of so-called \emph{probability-matching} algorithms for exploitation-exploration problems. Such algorithms are relatively less known. Recently, \citep{chapelle2011empirical} evaluates the performance of TS compared with other famous UCB-like algorithms \citep{carpentier2011finite}. And \citep{chapelle2011empirical} shows that TS has a \emph{state-of-the-art} performance in various tasks. After that many works on analyze TS appears \citep{kaufmann2012thompson,agrawal2012analysis,guha2014stochastic}. It turned out that TS has nice properties in a theoretical point of view. However, TS is still not very popular in practice. A possible reason is that inference the posterior is usually intractable. This paper tries to resolve the problem.

\textbf{Gumbel-Max trick}: Gumbel-Max trick is a tool to connect sampling problems and optimization problems, and has been used in various problems \citep{chen2016scalable,maddison2014sampling,papandreou2011perturb}. The most related work is \citep{chen2016scalable}, which also exploits the Gumber-Max trick for sampling a discrete random variable and relates to multi-armed bandits. But they consider the distributions with $P(x)\propto p_0(x)\prod_{i=1}^n p_i(x)$ which is significantly different with our problem.

\textbf{Best arm identification in fixed confidence setting}:  This setting comes from a fundamental question about exploration and exploitation tradeoff: when can an agent stop learning and start exploiting the learned knowledge? Many algorithms has been proposed. They mainly fall into two categories: The first one is LUCB \citep{kalyanakrishnan2012pac}, in which the agent pull arms according to the confidence bound; the second is the racing algorithms which is first introduced by \citep{maron1997racing}. In a racing algorithm, the agent maintains a set of active arms, and it pulls all active arms in each round and eliminate the suboptimal arms according to certain elimination rules.
In this paper, we only design a racing algorithm for our problem, since the racing algorithm is suitable for our problem. For more details, see Section \ref{sec:algorithm}.

\section{Preliminaries of Thompson Sampling}\label{sec:preliminaries}

Let $K$ denote the number of arms, and $\pi$ denote the prior distribution over $\vct\mu=\{\mu_1,\cdots,\mu_K\}$, where $\mu_i$ is the expectation of arm $i$'s reward. 
Suppose upto time step $t$, the agent has chosen action $i$ for $\tau_{i,t}$ times, and received rewards $X_i(t)=\{x_{i,1},\cdots,x_{i,\tau_{i,t}}\}$. Let $X(t)=\{X_1(t),\cdots,X_K(t)\}$ be all the observed rewards until time step $t$. 

{Thompson sampling} selects each arm randomly according to its (posterior) probability to be optimal, which is 
\begin{align}\label{eq:ts1}
\forall i\in [K],~~ P_i(t):=P\left(\mu_i=\max_{j}\mu_j | X(t) \right),
\end{align}
and $[K] := \{1, \cdots, K\}$ denotes the set of integers from 1 to $K$. 
Previous implementation \citep{chapelle2011empirical} of TS is outlined in Alg. \ref{alg:ts}, in which the Lines 6 and 7 essentially draw samples from $P_i(t)$. 
As we can see in Line \ref{line:posterior} of Alg. \ref{alg:ts}, this implementation requires to infer the posterior distribution of the mean rewards, which is efficient if the prior is conjugate.
However, in practice non-conjugate priors are more flexible in many situations such as where the arms are not independent, as discussed in the introduction.

For a non-conjugate prior, one possible solution is to approximate the intractable posterior with a sequential Monte Carlo (SMC) sampler, which can be done as follows: At each time step $t$, we maintain a set of weighted particles $\{(\xi_{t}^i,w_{t}^i)\}_{i=1}^N$, where $w_t^i$ is the weight of particle $\epsilon_t^i$. Initially, these particles are sampled from the prior distribution, i.e., $\xi_{1}^i\sim \pi$, and the weights are equal (e.g., the unit 1). When we observed $X_t$, we reweight $w_{t+1}^i$ according to the likelihood function, that is $w_{t+1}^i = w_{t}^i P(X_t|\xi_t^i)$. We use $P(x=\xi_t^i)=\frac{w_t^i}{\sum_{i'}w_t^{i'}}$ to approximate the posterior. Though straightforward, SMC has some shortcomings. For example, there is no standard way to choose the number of particles and when the number of observations growth up, most particles' relative weight is approaching zero~\cite{doucet2001introduction}, it makes SMC being an inefficient approximation of the posterior
. Our empirical results in Section \ref{sec:exp} further demonstrate that SMC is not sufficient; thereby calling for a new algorithm. 


\begin{algorithm}[t]
\caption{Thompson sampling}
\label{alg:ts}  
\begin{algorithmic}[1]
\STATE Input: Prior distribution $\pi$.
\STATE $t=0$.
\STATE Maintain sets: $X_i=\emptyset,\forall i\in [K], X=\{X_1,\cdots,X_K\}$.
\WHILE {$t<T$}
	\STATE $t=t+1$.
	\STATE Draw samples $\vct\mu \sim P(\vct\mu|X)$.\label{line:posterior}
    \STATE $I_t=\arg\max_i \mu_i$.
    \STATE Take action $I_t$, and receive reward $x_t$, $X_{I_t}=X_{I_t}\cup \{x_t\}$.
\ENDWHILE
\end{algorithmic}
\end{algorithm}

\section{Algorithm}\label{sec:algorithm}
We now present our algorithm. For clarity, we consider the case that $P(x|\mu_i)$ is a Bernoulli distribution for each arm $i$, for which a conjugate prior or other easy priors exist, and we consider the general non-conjugate prior $\pi$. 
As we shall see in Section \ref{sec:extend}, our method can be easily extended to many popular distributions. 

With Bayes' rule and straight-forward computations, 
we can rewrite the posterior distribution in Eq. (\ref{eq:ts1}) as follows:
\begin{align}
	P_i(t) &= \frac{1}{P(X(t))}\int \pi(\vct\mu) \mathbbm{1}\left[ \mu_i = \max_j \mu_j \right] \prod_{j=1}^{K}\prod_{s=1}^{\tau_{j,t}}P_j(x_{j,s}|\mu_j)d\vct \mu , \label{eq:ts2}
\end{align}
where $P(X(t))=\int \pi(\vct \mu) P(X(t)|\vct \mu) d\vct \mu$ is the marginal likelihood of the observations upto time $t$.

It is not easy to deal with the discrete distribution $P_i(t)$ in Eq.~(\ref{eq:ts2}) directly, especially when the prior is non-conjugate. One key step to derive our algorithm is that we can reformulate TS as an optimization problem via the Gumbel-Max trick~\cite{papandreou2011perturb}, as detailed in Section~\ref{sec:gumbel} and followed by the racing algorithm in Section~\ref{sec:racing}.

\subsection{Represent Thompson sampling as a bandit problem}\label{sec:gumbel}

Consider a general $K$-dimensional discrete distribution $P=\{P_1,\cdots,P_K\}$, instead of directly drawing samples from $P$, \textbf{Gumbel-Max} provides an alternative way, with which we first draw $K$ i.i.d samples $\{\epsilon_1,\cdots, \epsilon_K\}$ from the Gumbel$(0,1)$\footnote{If $\epsilon_i\sim$ Gumbel$(0,1)$, then $P(\epsilon=x)\propto e^{-(x+e^{-x})}$. Moreover, it is easy to sample from Gumbel$(0,1)$---simply draw $u$ from the uniform distribution $U[0,1]$ and set $\epsilon$ as $-\log(-\log u)$.} distribution, and 
then set $I=\arg\max_{i\in [K]} \epsilon_i + \log P_i$. It was shown that we have the samples from the target distribution, that is, $P(I=i)=P_i$~\citep{kuzmin2005optimum}. So, the Gumbel-Max trick provides a nice way to turn a sampling problem to an optimization problem. It has been used in various settings~\cite{chen2016scalable,maddison2014sampling,papandreou2011perturb}. 

Applying the Gumbel-Max trick to our problem in Eq. (\ref{eq:ts2}), we can represent TS as the following optimization problem: 
\begin{align}
I_t&=\argmax_{i\in [K]} \epsilon_i + \log P_i(t)\nonumber \\
=&\argmax_{i\in [K]} \log \int \pi(\vct\mu) e^{\epsilon_i}\mathbbm{1}\left[ \mu_i = \max_j \mu_j \right] \prod_{j=1}^{K}\prod_{s=1}^{\tau_{j,t}}P_j(x_{j,s}|\mu_j)d\vct \mu - \log P(X(t)), \nonumber  
\end{align}
where $\{\epsilon_i\}_{i=1}^K$ still denotes the set of samples from Gumbel$(0,1)$. However, it is still hard to directly solve it.

Our key idea to solve this problem efficiently is to construct a tractable distribution and further turn this problem as a best arm identification problem. Specifically, by introducing the conjugate prior, which is a Beta distribution for the Bernoulli case, we can reformulate the problem as follows:
\begin{align}
I_t=&\argmax_{i\in [K]}\log \int \pi(\mu)e^{\epsilon_i} \mathbbm{1}\left[ \mu_i = \max_j \mu_j \right]  \prod_{j=1}^{K}\frac{Z_j\cdot Beta(\mu_j|1,1)}{Z_j\cdot Beta(\mu_j|1,1)} \prod_{s=1}^{\tau_{j,t}}P_j(x_{j,s}|\mu_j)d\vct \mu - \log P(X) \nonumber \\
=&\argmax_{i\in [K]}\log \int \frac{e^{\epsilon_i}\pi(\vct\mu) \mathbbm{1}[\mu_i = \max_j \mu_j]}{\prod_{j=1}^{K}Beta\left( \mu_j|1,1\right)} \prod_{j=1}^K Beta\left( \mu_j | 1+o\left(X_j(t)\right), 1+z\left( X_j(t) \right) \right) d\vct \mu + C\nonumber  \\
=&\argmax_{i\in [K]}\log \mathbb{E}_{\forall j,\mu_j\sim Beta(\mu_j | 1+o(X_j(t)), 1+z(X_j(t)))}\left[ \frac{e^{\epsilon_i}\pi(\vct\mu) \mathbbm{1}[\mu_i = \max_j \mu_j]}{\prod_{j=1}^{K}Beta(\mu_j|1,1)} \right] + C\nonumber\\
=&\argmax_{i\in [K]} \mathbb{E}_{\forall j,\mu_j\sim Beta(\mu_j | 1+o(X_j(t)), 1+z(X_j(t)))}\left[ \frac{e^{\epsilon_i}\pi(\vct\mu) \mathbbm{1}[\mu_i = \max_j \mu_j]}{\prod_{j=1}^{K}Beta(\mu_j|1,1)} \right], \label{eq:gumbel}
\end{align}
where $Beta(\cdot | a, b)$ denotes the Beta distribution with parameters $(a,b)$, $o(X_j(t))$ and $z(X_j(t))$ represent the number of ones and the number of zeros in the observation sequence $X_j(t)$ respectively, $Z_j=\int Beta(\mu_j|1,1)\prod_{s=1}^{\tau_{j,t}}P_j(x_{j,s}|\mu_j)$ is a normalizer, 
and $C=-\log P(X(t))+\sum_{j}\log Z_j$ is a constant. In the above derivation, the second equality holds due to the conjugacy, and the last equality holds due to the fact that $\log (\cdot)$ is a monotonically increasing function. 

From the Gumbel-Max theory, we know that $I_t$ follows our target posterior distribution, that is, $I_t\sim P(t)$ \citep{papandreou2011perturb}. Essentially, the problem in Eq. (\ref{eq:gumbel}) is to find the variable with maximum expectation. As in our case, each variable corresponds to an arm, and this is known as a best arm identification (BAI) problem~\cite{jamieson2014best}. The benefit of our formulation is that we only need to draw samples from $Beta$ distribution, which can be done efficiently, in order to estimate the expectations. 

We solve the above BAI problem in the popular fixed-confidence setting~\cite{jamieson2014best}: for $\delta \in (0, 1)$ and $\sigma>0$, an algorithm is called $(\delta,\sigma)$-PAC \citep{kaufmann2015complexity} if and only if with probability at least $1-\delta$, it identifies arm $i$ such that $\mu_i > \max_{j\in[K]}\mu_j-\sigma$. This setting provides a simple and practical method to determine the number of samples we need to draw from the $Beta$ distribution: we can stop the sampling process if we are sure enough that we have identified a sufficiently good arm. 

\subsection{The racing algorithm}\label{sec:racing}

For the clarity of notations, let $f_i(\mu,t)=\frac{e^{\epsilon_i}\pi(\vct\mu) \mathbbm{1}[\mu_i = \max_j \mu_j]}{\prod_{j=1}^{K}Beta(\mu_j|1,1)}$ as the function within the expectation, $B_{j,t}(\cdot) = Beta(\cdot | 1+o(X_j(t)), 1+z(X_j(t))))$ as the Beta distribution for arm $j$ at time $t$, and $B_t(\vct \mu)=\{B_{1,t}(\mu_1),\cdots, B_{K,t}(\mu_K)\}$ as the collection of Beta distributions for all arms. 
We further use $v_i=f_i(\vct\mu,t)$ to denote a random variable, where $\vct\mu\sim B_t$. Recall that our goal is to identify the arm with the maximum expectation $I_t = \argmax_i \mathbb{E}[v_i]$. Suppose we have a set of samples $d=\{d_1,\cdots, d_m\}$ where $d_j\sim B_t$. We use $f_{i,m}=\frac{1}{m}\sum_{i=1}^m f_i(d,t)$ as our unbiased estimator of $\mathbb{E}[v_i]$.

Recall that the goal of \textbf{best arm identification} (BAI) is to identify the one with highest expectation among a set of random variables.
Following~\citep{kaufmann2015complexity}, 
a practical BAI 
algorithm in the fixed confidence setting typically consists of:
\begin{itemize}
\item {Policy}: given a sequence of past observations, a policy determines which arms to pull.
\item {Stopping rule}: a stopping rule can be described as a series of observation sets $\mathcal{F}_t,t\in \mathbb{N}_+$\footnote{$\mathbb{N}_+$ is the set of positive integers.}
, where $F_t$ is a set of observations. 
When an element $o\in \mathcal{F}_t$ is observed, the policy stops sampling.
\item {Recommendation rule}: 
a recommendation rule is usually to recommend the best arm. A BAI algorithm usually recommends the arm with the highest empirical mean.
\end{itemize}
Our algorithm imitates the racing algorithms \citep{maron1997racing,even2006action} for BAI problems. In racing algorithms, people maintain a set of arms as the candidates of the best arm. The policy of a racing algorithm is to draw a sample from the underlying distribution of each remained arm during each round. And then eliminate the suboptimal arms if the gap between the empirical means of the suboptimal arm and the maximum one if bigger than a threshold function. A racing algorithm stops if and only if only one arm is not eliminated.
The reason we choose racing algorithm is that we can compute $f_{i,m}$ with the same $d$, so drawing a sample from $B_t$ is essentially pulling all arms at the same time.
 Our algorithm is shown in Alg. \ref{alg:bandits}. In lines \ref{line:loop}-\ref{line:endloop} of Alg. \ref{alg:bandits}, we solve the BAI problem via a racing algorithm: we sample $\hat{\vct \mu}\sim B_t$ repeatedly until the empirically best arm $i_1$ is better than others $i_2$ significantly, e.g, larger than a threshold function defined by $\beta(m,\delta)$.
 
Theorem \ref{thm:delta-pac} guarantees that Alg. \ref{alg:bandits} is $(\delta,\sigma)$-PAC.
\begin{Thm} \label{thm:delta-pac}
If the threshold function $\beta(m,\delta)$ satisfies the following condition, 
\begin{align}
P\left(\exists m>0: |f_{i,m}-\mathbb{E}[v_i] |>\beta(m,\delta) \right)<\delta\label{eq:threshold}.
\end{align}
and $v_i$ is 1-subgaussian, then in Alg. \ref{alg:bandits}, at each time step $t$, with probability at least $1-K\delta$, $\mathbb{E}[v_{I_t}]>\mathbb{E}[\bar{f}_i] - \sigma$ for all $i$. 
\begin{proof}
We exploit standard arguments to prove the theorem. When Line \ref{line:break} is executed, and $I$ is a bad arm, that is $\mathbb{E}f_{I}<\mathbb{E}\bar{f}_i - \sigma$. By $\bar{f}_{I} - \bar{f}_{i^*}>2\beta(m,\delta) - \sigma$. It is easy to see that at least one of two following events happen: $\bar{f}_I - \beta(m,\delta)>\mathbb{E}f_I$ or $\bar{f}_{i^*}+\beta(m,\delta)-\sigma < \mathbb{E}f_{i^*}-\sigma$. With Eq. (\ref{eq:threshold}) and union bound, we complete the proof.
\end{proof}
\end{Thm}

As stated in Theorem \ref{thm:delta-pac}, our algorithm requires the threshold function $\beta(m,\delta)$ to satisfy the condition in InEq. (\ref{eq:threshold}). We choose the threshold function $\beta(m,\delta)$ as stated in Eq. (\ref{eq:beta}) and Theorem 8 in \citep{kaufmann2015complexity} shows the chosen threshold function satisfies InEq. (\ref{eq:threshold}).

\textbf{Threshold function\footnote{This function works for $1$-subgaussian random variables. In practice, we can rescale $v_i$ to make it $1$-subgaussian.} ( \citep{kaufmann2015complexity})}: 
\begin{align}
\beta(m, \delta)=\sqrt{\left( \log \frac{1}{\delta} + 3\log\log\frac{1}{\delta} + \frac{3}{2} \log \log \frac{em}{2} \right) / (2m)} . \label{eq:beta}
\end{align}

It is noteworthy that a common assumption in the literature of multi-armed bandits is that the underlying random variable is subgaussian. This is because previous works are relying on Chernoff-hoeffding's inequality which makes the assumption. We make the same assumption in this work, that is $v_i$ is subgaussian, and leave the excluded cases to future work. 

\begin{algorithm}[t]
\caption{Racing Thompson}
\label{alg:bandits}  
\begin{algorithmic}[1]
\STATE Input: Prior distribution $\pi$, parameters $\delta, \sigma$, and time horizon $T$.
\STATE $t=0$.
\STATE $X_i=\emptyset,\forall i\in [K], X=\{X_1,\cdots,X_K\}$.
\WHILE {$t<T$}
	\STATE Draw $K$ i.i.d samples $\epsilon_i$ from the Gumbel$(0,1)$ distribution. 
	\STATE $m=1,d=\emptyset$.
    \LOOP \label{line:loop}
    	\STATE Draw a sample $\hat{\vct \mu} \sim B_t$, $d=d\cup\{\hat{\vct\mu}\}$.
        \STATE $m=m+1$.
        \STATE $\bar{f}_i=\frac{1}{m}\sum_{j=1}^m f_i(d_j,t)$. 
        \STATE Identify the best arm $i_1=\arg\max_{i\in [K]}\bar{f}_i$ and the second best $i_2=\arg\max_{i\in [K]\backslash \{i_1\}}\bar{f}_i$.
        \IF {$\bar{f}_{i_1} - \bar{f}_{i_2}>2\beta(m,\delta) - \sigma$}
        \STATE Break loop.\label{line:break}
        \ENDIF
    \ENDLOOP\label{line:endloop}
    \STATE $I_t=\arg\max_i \frac{1}{m}\sum_{j=1}^m f_i(d_j,t)$.
    \STATE Take action $I_t$, and receive reward $x_t$, $X_{I_t}=X_{I_t}\cup \{x_t\}$.
\ENDWHILE
\end{algorithmic}
\end{algorithm}

\subsection{Extend to other distributions}\label{sec:extend}
In Section \ref{sec:gumbel}, we use Bernoulli distribution as an example to introduce our algorithm. In the derivation of Eq. (\ref{eq:gumbel}), we can see that the central step is on constructing a prior distribution whose posterior is computationally tractable. In the case of Bernoulli distribution, we choose Beta distribution which is the conjugate prior. For many widely-used distributions, there exist such priors. For example, any exponential family distribution exists a conjugate prior. 
And the posterior inference for many of them are tractable, see \citep{george1993conjugate} for more details.
Beyond exponential family distributions with conjugate priors, there are tractable distributions with other priors. For example, many 1-dimensional exponential family distributions with a non-informative Jeffreys prior is tractable \citep{jaynes1968prior}. The Jeffreys prior is proportional to the square root of the Fisher information matrix and maybe improper (e.g., the Jeffreys prior for Gaussian is uniform in an infinite space). Some tractable and representative exponential family distributions with Jeffreys prior are listed in Table \ref{table:distributions}.

\begin{table}[ht!]
\begin{center}
	\begin{tabular}{c | c | c | c}
	    \hline
     	Name & distribution &  prior & posterior\\
        Bernoulli & $p^x(1-p)^{1-x}$ & $Beta(1,1)$ &  $Beta(1+s,1+T-s)$\\
        Exponential & $\lambda e^{-\lambda x}$&$\Gamma(1,1)$&$\Gamma(1+T,1+s)$\\
        Gaussian &$\frac{1}{\sqrt{2\pi}}\exp\{-\frac{(x-\mu)^2}{2}\}$&$\propto 1$&$\mathcal{N}(\frac{s}{n},\frac{1}{n})$\\
        Poisson &$\frac{\lambda^xe^{-\lambda}}{x!}$&$\propto\frac{1}{\sqrt{\lambda}}$&$\Gamma(\frac{1}{2}+s, T)$\\
        \hline
	\end{tabular}
\end{center}
\caption{Examples of distributions with Jeffreys prior. $s$ is the sum of $T$ observations.}
\label{table:distributions}
\end{table}

\section{Experiments}
\label{sec:exp}
We do experiments on Bernoulli bandits, i.e. $P(x|\mu_i)$ is a Bernoulli distribution. Our racing algorithm has two parameters $\delta, \sigma$ as shown in Alg. \ref{alg:bandits}, where these parameters balance the number of samples we need to draw and the accuracy of Alg. \ref{alg:bandits} (Please see Section \ref{sec:racing}). 
We first empirically analyze Alg \ref{alg:bandits}'s sensitivity to parameters $\delta, \sigma$. We use $Beta(5,5)$ as the prior and set time steps $t=1000$. We repeat the experiment 10 times and present the average results in 
Table \ref{table:sensitivity} where the first number in each entry is the average number of particles, and the second number is the average regret. Table \ref{table:sensitivity} shows that the smaller the parameters $\delta, \sigma$ are, the more particles we use, and the regret is smaller. This is consistent with the definition of $\delta$ and $\sigma$ (Please see Section \ref{sec:gumbel}).
\begin{table}[ht!]
\caption{Empirical evaluation of sensitivity on parameters $\delta, \sigma$.  The first number in each entry is the average number of particles, and the second number is the average regret. }
\begin{center}
	\begin{tabular}{c| c c c c}
     \hline
     	& $\sigma=0.1$ & $\sigma=0.3$ & $\sigma=0.5$ & $\sigma=0.7$ \\
        \hline
        $\delta=0.1$ & $353.4/19.3$& $36.4/21.6$ & $13.4/23.2$ & $6.3/27.8$ \\
        $\delta=0.3$ & $251.6/19.8$ & $26.7 / 21.4$ & $8.9 / 24.8$ & $4.8 / 28.0$\\
        $\delta=0.5$ & $213.0 / 20.1$ & $ 20.3 / 21.8$ & $7.2 / 24.5$ & $2.9 / 29.2$ \\
        $\delta=0.7$ & $184.9 / 20.2$ & $17.9/ 22.8$ & $5.5 / 25.2$ & $2.0 / 29.4 $\\ 
        \hline
	\end{tabular}
\end{center}
\label{table:sensitivity}
\end{table}

We set $\delta$ and $\sigma$ to be $0.1$ and $0.1$ respectively in following experiments. 
Recall that the agent selects arm $I_t$ at time step $t$, and the goal is to maximize the cumulative reward $\sum_{t=1}^{T}\mu_{I_t}$, where $\mu_i$ is the unknown mean of arm $i$, and $\vct\mu\sim \pi$. It is easy to see that maximizing the cumulative reward is equivalent to minimizing the regret:
\begin{align}
T \max_{i\in [K]} \mu_i - \sum_{t=1}^{T}\mu_{I_t}\label{eq:regret}.
\end{align} 
In the first experiment, we follow the setting in \citep{chapelle2011empirical} and compare the regret between Alg. \ref{alg:ts} and Alg. \ref{alg:bandits}. To make Alg. \ref{alg:ts} computationally efficient, we use the Beta distribution which is the conjugate prior so that the standard TS can apply. There are $10$ arms, and the prior of the $i$-th arm follows $Beta(\cdot | a,b)$, where $a$ and $b$ are uniformly selected
from the interval $(1.0, 10.0)$. We repeat 100 times and present the average results in Figure 1(a). 
From Figure 1(a), we can see that our racing Alg.~\ref{alg:bandits} has a competitive performance with the vanilla Thompson sampling (i.e., Alg. \ref{alg:ts}).
And on average, Alg. \ref{alg:bandits} uses about $400$ samples in each time step.

\begin{wrapfigure}{r}{0.50\linewidth}
  \centering
    \subcaptionbox{Conjugate prior\label{fig:vi-mixture}}
    {\includegraphics[height=3cm]{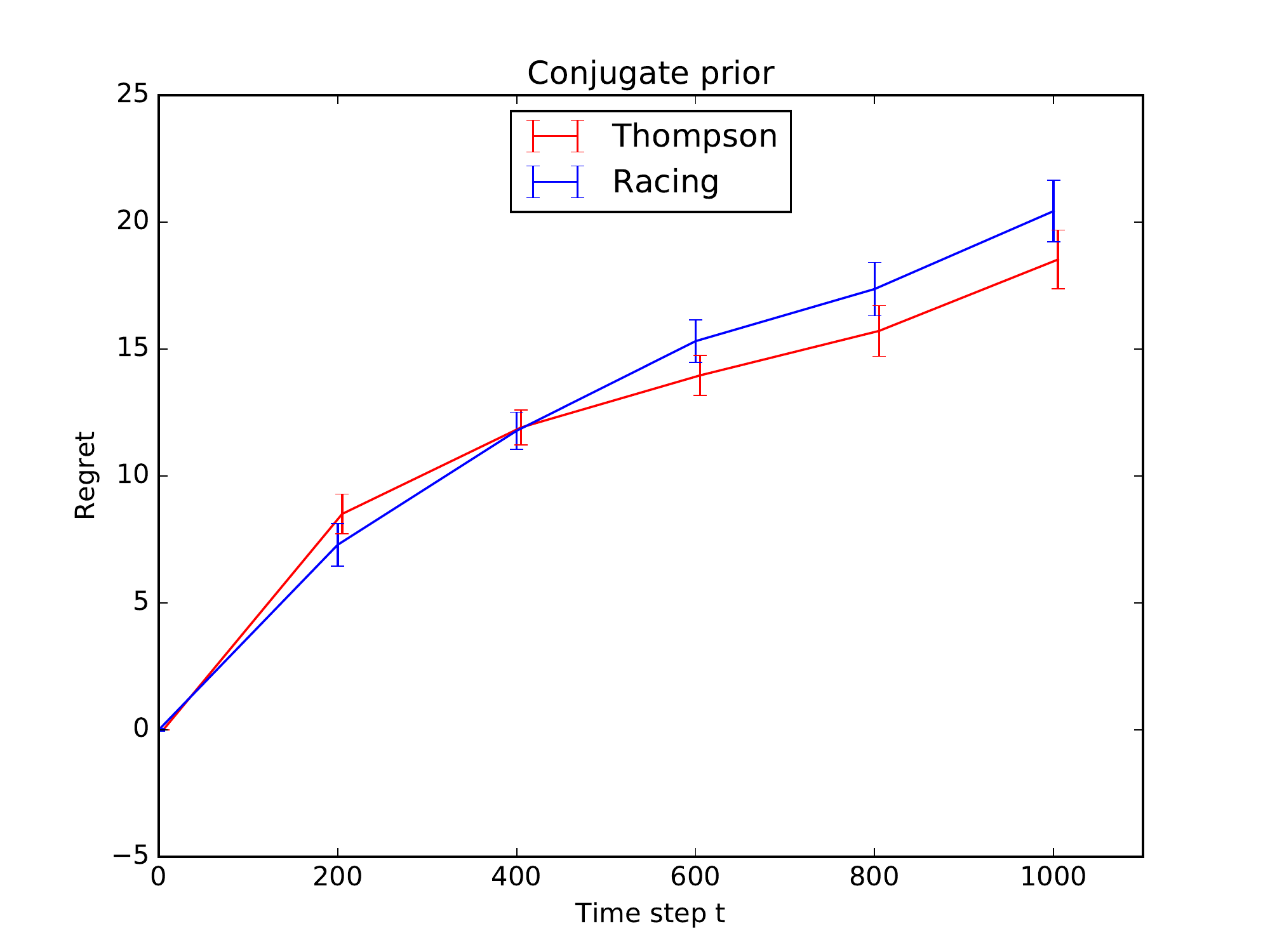}}
   \subcaptionbox{Non-conjugate prior\label{fig:kdrf-mixture}}
     {\includegraphics[height=3cm]{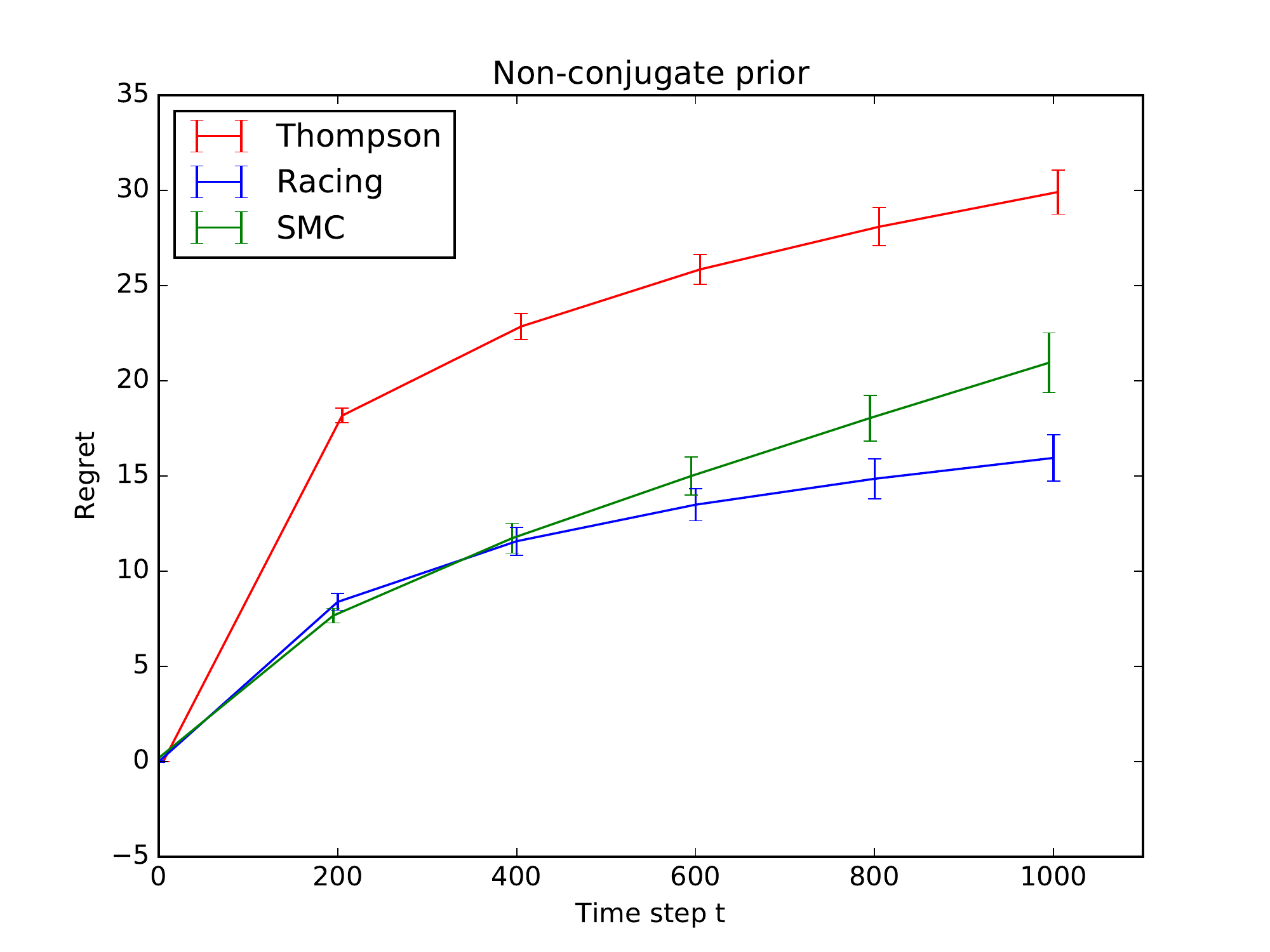}}
        \caption{The regret of Bernoulli bandits with ({\bf a}) a conjugate prior and  ({\bf b}) a non-conjugate prior.\label{fig:conjugate}}
\end{wrapfigure}

We evaluate the performance of Alg. \ref{alg:bandits} for a non-conjugate prior in the second experiment. There are $10$ arms, and the prior of the $i$-th arm is a truncated Gaussian distribution where the truncated interval is $[0,1]$. Since a Gaussian distribution is not a conjugate prior of the Bernoulli distribution, posterior inference is computationally inefficient.
We implement Alg.~\ref{alg:ts} in two settings: 
\begin{enumerate}
\item {\bf SMC}: the first is to use sequential Monte-Carlo (SMC) to approximate the posterior. 
To make the comparison fairly, we use the same the number of particles as used in Alg. \ref{alg:bandits} at the corresponding time step.
\item {\bf Thompson}: the second method is to use a $Beta$ distribution as (improper) prior so that we can infer the posterior efficiently in closed-form and the standard TS applies.
\end{enumerate}
We present the results in Figure 1(b), where we can see that Alg.~\ref{alg:bandits} outperforms the second variant of Alg. \ref{alg:ts} (i.e., Thompson) significantly.  

When time step $T$ is relatively small, our algorithm has a similar performance with the first variant of Alg. \ref{alg:ts} (i.e., SMC), and as $T$ grows up, our algorithm outperforms SMC.
The reason is that by Bayes' rule, we have $P(\vct\mu | X)\propto \pi(\vct \mu)\prod_{i=1}^n P(X_i | \vct \mu)$ where $n$ is the number of observations. When $n$ grows up, likelihood $\prod_{i=1}^n P(X_i|\vct \mu)$ "dominates" the posterior. When we use SMC to approximate the posterior, the prior distribution determines $\xi_t^i$ (See Section \ref{sec:preliminaries}.), and likelihood determines $w_t^i$. According to concentration inequality, almost all particles' relative weight will grow to $0$\footnote{This is an informal discussion, see \citep{doucet2001introduction} for more detailed discussions about the shortcomings of SMC.}, this fact makes SMC cannot fit the posterior well. On the other hand, in our algorithm, the particles (Set $d$ in Alg. \ref{alg:bandits}) are sampled from a constructed posterior, which can fit the likelihood function well. This is the reason why Alg. \ref{alg:bandits} outperforms Alg. \ref{alg:ts} with SMC significantly.

\begin{wrapfigure}{r}{0.5\textwidth}\label{fig:dependent} \vspace{-.4cm}
\begin{center}
\includegraphics[width=0.5\textwidth]{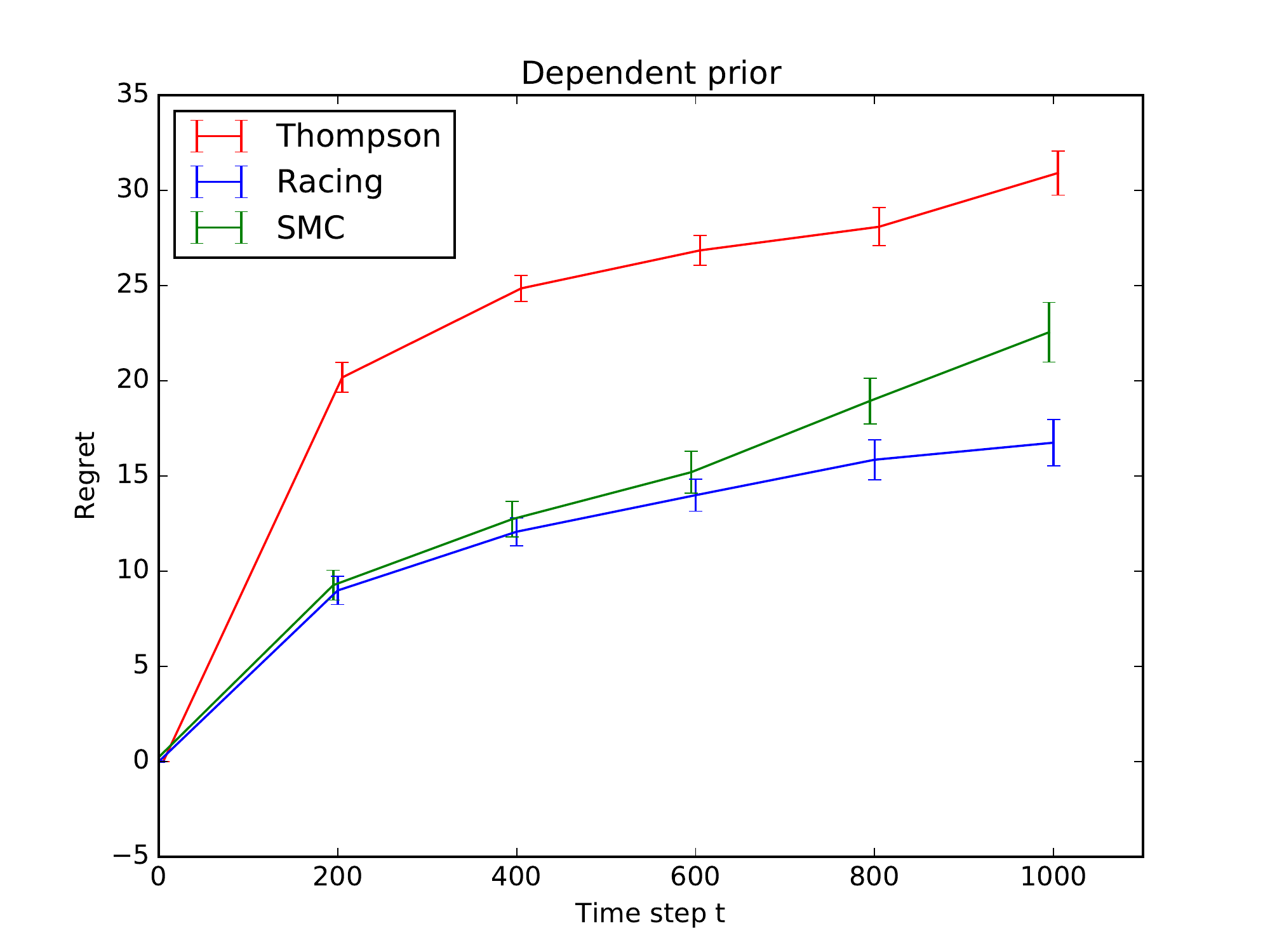}
\end{center}
\caption{The regret of Bernoulli bandits with a dependent prior.}
\end{wrapfigure}

Our last experiment is on structured priors, in which the arms may not be independent. In real-world applications, arms are usually dependent. For example, if we want to recommend articles to users, we can view each article as an arm. A natural way is to first analyze the topics of articles via topic models \citep{blei2003latent}, and then build a prior distribution according to the topics. So such priors are not independent and non-conjugate. We also do experiments on synthetic data with 10 arms. Suppose there is a real-valued vector $u_i$ associated with each arm, and suppose the $L_2$-norm of $u_i$ is $1$. These vectors can be interpreted as topic vectors (the proportion of each topic), in topic models or some other compact representation of arms. For convenience, suppose the prior is a 10-dimensional Gaussian distribution with each dimension's mean is $0.5$ and we restrict the value within $[0,1]$. We define the covariance matrix $\Sigma$ of the prior distribution as follows: $\Sigma_{i,i}=1$ for all $i$ and $\Sigma_{i,j}=u_i \cdot u_j$, where $x\cdot y$ denotes the inner product between vectors $x$ and y. We use the same baselines (i.e., SMC and Thompson) as in the previous experiment. We present the results in Figure 2. Again, we can see that our racing algorithm outperforms the baselines as in the previous experiment, namely, our racing algorithm significantly outperforms Thompson which adopts an improper prior, and our method outperforms SMC when the time step $t$ is relatively large. 

\section{Conclusions and discussions}\label{sec:conclusions}
We present an efficient racing algorithm for Thompson sampling with general non-conjugate priors. Our method is built on a new reformulation of Thompson sampling as a best arm identification problem based on the Gumbel-Max trick. 
Our racing algorithm has a theoretical guarantee and works well empirically for both conjugate and non-conjugate priors. 

However, there are several open problems. As we mentioned in Section \ref{sec:racing}, our stopping rule requires that the random variables are subgaussian. It is natural to ask that how to identify the best random variable if the variables are not subgaussian? The second one is that we construct a tractable proposal distribution and draw samples from it instead of inferring the posterior. So is there a unified framework to construct such proposal distributions if $P(\cdot|\mu_i)$ has some properties, such as $P(\cdot|\mu_i)$ is subgaussian? We would like to see future work on these problems.
\subsubsection*{Acknowledgments}
The work was supported by the National Basic Research Program (973 Program) of China (No. 2013CB329403), NSFC Projects (Nos. 61620106010, 61621136008), and the Youth Top-notch Talent Support Program.

\bibliographystyle{plain}

\small
\bibliography{ref}

\end{document}